\documentclass{article}

\usepackage{arxiv}

\usepackage[utf8]{inputenc} 
\usepackage[T1]{fontenc}    
\usepackage{hyperref}       
\usepackage{url}            
\usepackage{booktabs}       
\usepackage{amsfonts}       
\usepackage{nicefrac}       
\usepackage{microtype}      
\usepackage{lipsum}
\usepackage{textcomp} 
\usepackage{graphicx}
\graphicspath{ {./images/} }
\usepackage{amsmath}
\usepackage{amssymb}
\DeclareUnicodeCharacter{223C}{$\sim$}
\newcommand{\cev}[1]{\reflectbox{\ensuremath{\vec{\reflectbox{\ensuremath{#1}}}}}}
\usepackage{booktabs}
\usepackage{subcaption}
\usepackage{microtype}
\usepackage{xurl}
\usepackage{lineno}
\usepackage{gensymb}
\usepackage{color}

\title{Efficient Kilometer-Scale Precipitation Downscaling with Conditional Wavelet Diffusion}

\author{
 Chugang Yi \\
 Department of Mathematics \\
 University of Maryland \\
 College Park, MD \\
 \And
 Minghan Yu \\
 Department of Mathematics \\
 University of Maryland \\
 College Park, MD \\
 \And
 Weikang Qian \\
 Department of Geography \\
 University of Florida \\
 Gainesville, FL, USA \\
 \And
 Yixin Wen \\
 Department of Geography \\
 University of Florida \\
 Gainesville, FL, USA \\
 \And
 Haizhao Yang \\
 Department of Mathematics \\
 Department of Computer Science \\
 University of Maryland \\
 College Park, MD \\
}

\begin{document}
\maketitle
\begin{abstract}
Effective hydrological modeling and extreme weather analysis demand precipitation data at a kilometer-scale resolution, which is significantly finer than the 10 km scale offered by standard global products like IMERG. To address this, we propose the Wavelet Diffusion Model (WDM), a generative framework that achieves 10x spatial super-resolution (downscaling to 1 km) and delivers a 9x inference speedup over pixel-based diffusion models. WDM is a conditional diffusion model that learns the learns the complex structure of precipitation from MRMS radar data directly in the wavelet domain. By focusing on high-frequency wavelet coefficients, it generates exceptionally realistic and detailed 1-km precipitation fields. This wavelet-based approach produces visually superior results with fewer artifacts than pixel-space models, and delivers a significant gains in sampling efficiency. Our results demonstrate that WDM provides a robust solution to the dual challenges of accuracy and speed in geoscience super-resolution, paving the way for more reliable hydrological forecasts.
\end{abstract}

\section{Introduction}

Accurate, high-resolution precipitation data are critical for flood forecasting, hydrologic modeling, infrastructure design, and disaster response \cite{adler2012estimating,BehrangiWen2017,behrangi2012quantification,Cerveny2007} . The inherent strong spatio-temporal variability of precipitation poses substantial modeling challenges, as unresolved small-scale fluctuations can propagate into large uncertainties in crucial outputs such as runoff calculations and flood inundation maps \cite{BehrangiWen2017,Wen2021,rs12081258}. Consequently, the reliability of precipitation estimates is fundamentally tied to the observing system's capacity to resolve these fine-scale features. While ideal sensors would offer continuous, high-resolution observations at kilometer or even sub-kilometer scales, such comprehensive coverage is often impeded by the economic and technical constraints of deploying dense observational networks (e.g., radar, in-situ gauges), particularly in data-sparse regions \cite{behrangi2024comparative}. Global satellite-derived products, such as those from the Global Precipitation Measurement (GPM) mission's Integrated Multi-satellitE Retrievals (IMERG) algorithm, represent a significant advancement by providing consistent global precipitation estimates (e.g., at ∼10 km resolution and 30-minute intervals) \cite{schneider2013climate,HuffmanTanNCAR:GPMIMERG:2025}. However, this resolution level is insufficient to capture localized extremes, such as intense convective rainfall in mountainous or urban environments, where variability often occurs at scales less than 10 km \cite{Mazzoglio2019}. The resolution gap between global data like IMERG \cite{HuffmanTanNCAR:GPMIMERG:2025} and the needs of local applications necessitates the development of effective downscaling methodologies.

In this paper, our goal of downscaling is to transform coarse-resolution inputs (e.g. 10 km) into observational high-resolution fields (e.g., 1 km) that better capture the spatial variability of precipitation, especially in regions with complex topography or intense convective activity. Traditional statistical downscaling methods such as regression-based models, include bias correction with spatial disaggregation, and kriging \cite{wilby1997downscaling,wilby1998statistical,tareghian2013statistical}. While these methods are computationally efficient and interpretable, these methods are primarily designed for linear and stationary scenarios, which limits their effectiveness when applied to a complex, non-linear system like precipitation. Traditional dynamical downscaling methods—using regional climate models (RCMs) or Earth system models (ESMs)—numerically solve governing equations at higher resolutions, preserving physical consistency through explicit fluid dynamics and thermodynamics. These methods offer greater physical consistency but are computationally intensive, making them impractical for many operational or real-time applications \cite{Fowler2007, Wood2004, Hong2014,Glotter2014,Wang2021,dai2025precipdiff}. 

To overcome the previously mentioned limitations, deep generative models such as generative adversarial network \cite{ledig2017photo}, variational autoencoder \cite{Harris2022} and normalizing flow \cite{lugmayr2020srflow} offer a promising new avenue for statistical downscaling. Among these, diffusion models (DMs) have proven especially powerful, achieving state-of-the-art results in high-fidelity image generation \cite{ho2020denoising, song2020score, dhariwal2021diffusion}. DMs are also well-suited to geoscience applications due to their ability to capture complex geographic patterns, integrate heterogeneous data sources (e.g., satellite, radar, reanalysis), and generate ensembles for uncertainty quantification \cite{Ling2024, Harris2022, Harder2023}. In downsacling tasks, these models are adapted to generate high-resolution fields conditioned on the coarser inputs \cite{lopezgomez2024dynamicalgenerativedownscalingclimatemodel, watt2024generativediffusionbaseddownscalingclimate, Mardani2025}, and have shown strong results, particularly for smooth variables such as temperature or reanalysis data like ERA5 \cite{bell2021era5}.

Precipitation, in particular, is characterized by high intermittency, sharp gradients, complex multi-scale structures, and strongly non-Gaussian behavior\cite{lussana2022wavelet,kumar1994wavelet}. While reanalysis products like ERA5 provide valuable global coverage, their spatial smoothing \cite{bell2021era5,hersbach2020era5,simmons2020global} leads to the loss of critical fine-scale details and an underestimation of localized, high-intensity events, which are better preserved in direct observational data such as the Multi-Radar Multi-Sensor (MRMS) product \cite{zhang2016multi}. Our work specifically addresses the challenge of generating high-fidelity, kilometer-scale precipitation fields consistent with observations, using the MRMS dataset for training and evaluation. However, accurately representing precipitation at high spatial resolutions presents substantial modeling difficulties. Many contemporary diffusion models typically operate in the pixel domain \cite{karras2022elucidating,watt2024generativediffusionbaseddownscalingclimate,Mardani2025, liu2024generative,lopezgomez2024dynamicalgenerativedownscalingclimatemodel}. This makes it inherently challenging for them to effectively learn and reproduce the complex, multi-scale dependencies and sharp features of precipitation, often resulting in overly smooth or artifact-laden outputs. 

To address this critical gap, we introduce a novel Wavelet Diffusion Model (WDM). WDM is specifically designed to overcome the limitations of pixel-space models by operating entirely in the wavelet domain. The core motivation for this approach lies in the inherent capabilities of wavelet transforms, which are particularly well-suited for representing the complex, multi-scale structures found in precipitation fields \cite{lussana2022wavelet,kumar1994wavelet}. Wavelet transforms perform a Multi-Resolution Analysis (MRA), decomposing signals into different scale components. This allows for the explicit representation and modeling of the multi-scale organization of precipitation, from broad patterns to localized elements, as demonstrated in studies using wavelets to analyze precipitation structures across various spatial scales \cite{phung2023wavelet,jiang2023low}. This localization also enables the efficient representation of sharp gradients and discontinuities, which are key features of precipitation. By leveraging the wavelet domain, WDM significantly improves the quality of generated high-resolution precipitation fields from observational data, as well as the efficiency of inference in the diffusion sampling process.

The primary contributions of this work are summarized as follows:
\begin{enumerate}
    \item We propose a novel wavelet diffusion model, a conditional generative model that performs downscaling in the wavelet domain. This approach is specifically designed to capture complex spatial structures more effectively than pixel-space methods.
    \item We apply the proposed model to achieve 1-km resolution precipitation downscaling using \emph{observational} MRMS radar data instead of model-based ERA5 reanalysis data. To the best of our knowledge, this is among the first successful applications of diffusion models to kilometer-scale precipitation generation.
    \item We demonstrate — both quantitatively and visually — that our new model, with significantly lower computational cost, outperforms pixel-based diffusion models in capturing fine spatial detail, reducing artifacts, and achieving better alignment with evaluation metrics.
\end{enumerate}

The structure of this paper is organized as follows. The methodologies of this paper are introduced in Section \ref{sec:method}, including the formulation of our wavelet diffusion model. The datasets utilized in the numerical examples are described in Section \ref{sec:data}, including the MRMS high-resolution precipitation fields and the data preprocessing procedures. Comprehensive experimental results are presented in Section \ref{sec:results}, including quantitative evaluations, spectral analyses, and meteorological skill assessments. The implementation details of our numerical examples are discussed in Appendix \ref{sec:implementation}. Discussions on the implications, limitations, and future research directions are provided in Section \ref{sec:conclusion}.

\section{Methodology}\label{sec:method}
Our methodology addresses the task of precipitation downscaling by learning the conditional distribution $p(\mathbf{x}^{\text{high}} \mid \mathbf{x}^{\text{low}})$, where $\mathbf{x}^{\text{high}}$ is a high-resolution precipitation field (target) and $\mathbf{x}^{\text{low}}$ is its low-resolution observation (condition). To this end, we propose the Wavelet Diffusion Model (WDM) that integrates the generative power of diffusion models with the wavelet transform's ability to disentangle low-frequency patterns from its unique, high-frequency directional characteristics. In this section, we will first review the background of conditional diffusion models, then introduce the wavelet decomposition as a powerful representation for meteorological data, and finally present the complete WDM framework for high-fidelity downscaling.
\subsection{Diffusion Models}
We first introduce the general setting in the diffusion model using the variance preserving stochastic differential equation (VP-SDE) formulation from \cite{song2020score}.

\textbf{The Forward and Reverse Process.} Let $\mathbf{x}_0 \sim p_0(\mathbf{x})$ denotes samples from a general target distribution. The forward process $\mathbf{x}_t$ progressively injects Gaussian noise into target samples over time $t$, which is governed by the SDE:
\begin{equation}
d\mathbf{x}_t = -\tfrac{1}{2}\beta(t)\mathbf{x}_tdt +  \sqrt{\beta(t)}d\mathbf{W}_t,
  \label{eq:forward-sde}
\end{equation}
where $\mathbf{W}_t$ is a Wiener process, and $\beta(t)>0$ is a predefined schedule. Let $p_{t}(\mathbf{x)}$ be the probability distribution function of $\mathbf{x}_t$, $p_{t}(\mathbf{x)}$ will converge to $p_{\infty}(\mathbf{x}) = \mathcal{N}(0, \mathbf{I})$ as $t \to \infty$.

The corresponding reverse process, as derived in \cite{anderson1982reverse, song2020score}, provides a mechanism to transform the Gaussian noise back into samples from the target distribution:
\begin{equation}
d\cev{\mathbf{x}}_t = \Bigl[ -\tfrac{1}{2}\beta(T-t)\cev{\mathbf{x}}_{t} - \beta(T-t)\nabla_{\mathbf{x}} \log p_{T-t}(\cev{\mathbf{x}}_{t})\Bigr]dt + \sqrt{\beta(T-t)}d\bar{\mathbf{W}_t},  \quad t \in [0,T]
\label{eq:reverse-sde-general}
\end{equation}
where $\bar{\mathbf{W}}_t$ is a reverse-time Wiener process, $\cev{\mathbf{x}}_{T-t}=\mathbf{x}_t$ and $\cev{\mathbf{x}}_{0} \sim p_T(\mathbf{x})$. When $T$ is sufficiently large,  we can initialize the backward process~\eqref{eq:reverse-sde-general} with $\cev{\mathbf{x}}_{0} \sim \mathcal{N}(0, \mathbf{I})$ to obtain samples from the target distribution $p_0(\mathbf{x})$. The score function, defined as $S(\mathbf{x},t) := \nabla_{\mathbf{x}} \log p_t(\mathbf{x})$, is typically achieved by training a neural network $S_\theta(\mathbf{x},t)$ with parameters $\theta$ by minimizing the following score-matching loss \cite{vincent2011connection,song2020score,ho2020denoising}: 
\begin{equation}
\label{eq:score_matching}
  L_{\mathrm{score}}(\theta)
  :=
  \mathbb{E}_{t} \big\{w(t)
  \mathbb{E}_{\mathbf{x}_0 \sim \hat{p}_0}
  \;\mathbb{E}_{\mathbf{x}_t \sim p_t(\cdot \mid \mathbf{x}_0)}
    \big [ \bigl\|\,S_{\theta}\bigl(\mathbf{x}_t,t\bigr) -
    \nabla_\mathbf{x} \log p_t\bigl(\mathbf{x}_t \mid \mathbf{x}_0\bigr)
    \bigr\|_2^2 \big ] \big \},
\end{equation}
where $t\sim \mathcal{U} [0,T]$, $ p_t(\cdot \mid \mathbf{x}_0)$ is the distribution of $\mathbf{x}_t$ in the forward SDE (Equation~\eqref{eq:forward-sde}) initialized at $\mathbf{x}_0 $, $\hat{p}_0$ is empirical target distribution and $w(t)$ is a prescribed positive weighting function. Using the trained score function $S_{\theta}(\mathbf{x}_t,t)\approx \nabla_{\mathbf{x}} \log p_t(\mathbf{x})$, numerical solvers for SDEs (e.g., Euler-Maruyama) can be employed to simulate the reverse SDE in Equation~\eqref{eq:reverse-sde-general}, which can generate samples of the target distribution $p_0(\mathbf{x})$ from Gaussian noise. 

\textbf{Conditional Diffusion Model for Dowscaling.} For downscaling tasks, given low-resolution data $\mathbf{x}^{\text{low}}$, our goal is to generate corresponding high-resolution samples $\mathbf{x}^{\text{high}}$ from a learned conditional probability distribution $p_0^{\text{high}}(\cdot \mid \mathbf{x}^{\text{low}})$. The central strategy of the score-based generative model \cite{song2020score} is to reframe the problem of learning a target distribution into the task of learning its score function. As an analog of Equation~\eqref{eq:score_matching}, we parametrized the conditional score function $\nabla_\mathbf{x} \log p_{t}(\mathbf{x}_t^{\text{high}}\mid \mathbf{x}^{\text{low}})$ by a neural network $S_{\theta}\bigl(\mathbf{x}_t^{\text{high}},\,\mathbf{x}^{\text{low}},\,t\bigr)$, and trained by:
\begin{equation}
  L_{\mathrm{cond}} (\theta)
  :=
  \mathbb{E}_{t} \big\{w(t)
  \mathbb{E}_{(\mathbf{x}_0^{\text{high}}, \mathbf{x}^{\text{low}})}
  \;\mathbb{E}_{\mathbf{x}_t^{\text{high}}}
    \big [ \bigl\|\,S_{\theta}\bigl(\mathbf{x}_t^{\text{high}},\mathbf{x}^{\text{low}},t \bigr) -
    \nabla_\mathbf{x} \log p_t\bigl(\mathbf{x}_t^{\text{high}}\!\mid\!\mathbf{x}_0^{\text{high}}\bigr)
    \bigr\|_2^2 \big ] \big \},
    \label{eq:conditional_loss}
\end{equation}
where $(\mathbf{x}_0^{\text{high}}, \mathbf{x}^{\text{low}}) $ is sampled from the empirical joint distribution $\hat{p}_0(\mathbf{x}^{\text{high}},\mathbf{x}^{\text{low}})$, $\mathbf{x}_t^{\text{high}}\sim p_t(\cdot\mid \mathbf{x}_0^{\text{high}})$, and $p_t(\cdot\mid \mathbf{x}_0^{\text{high}})$ still comes from the same SDE (Equation~\eqref{eq:forward-sde}) for any initialization at $\mathbf{x}^{\text{high}}_0$. The minimizer of Equation~\eqref{eq:conditional_loss} is the true underlying conditional score function $\nabla_\mathbf{x} \log p_{t}(\mathbf{x}_t^{\text{high}} \mid \mathbf{x}^{\text{low}})$ when given sufficient data \cite{song2020score}. 
To sample from $p_0^{\text{high}}(\cdot \mid \mathbf{x}^{\text{low}})$, we can directly replace the unconditional score function in Equation~\eqref{eq:reverse-sde-general} with $S_{\theta}\bigl(\mathbf{x}_t^{\text{high}},\,\mathbf{x}^{\text{low}},\,t\bigr)$, and solve the $\cev{\mathbf{x}}_T$ for target samples as before. 
\begin{figure}[h!]
\centering
\includegraphics[width = \textwidth]{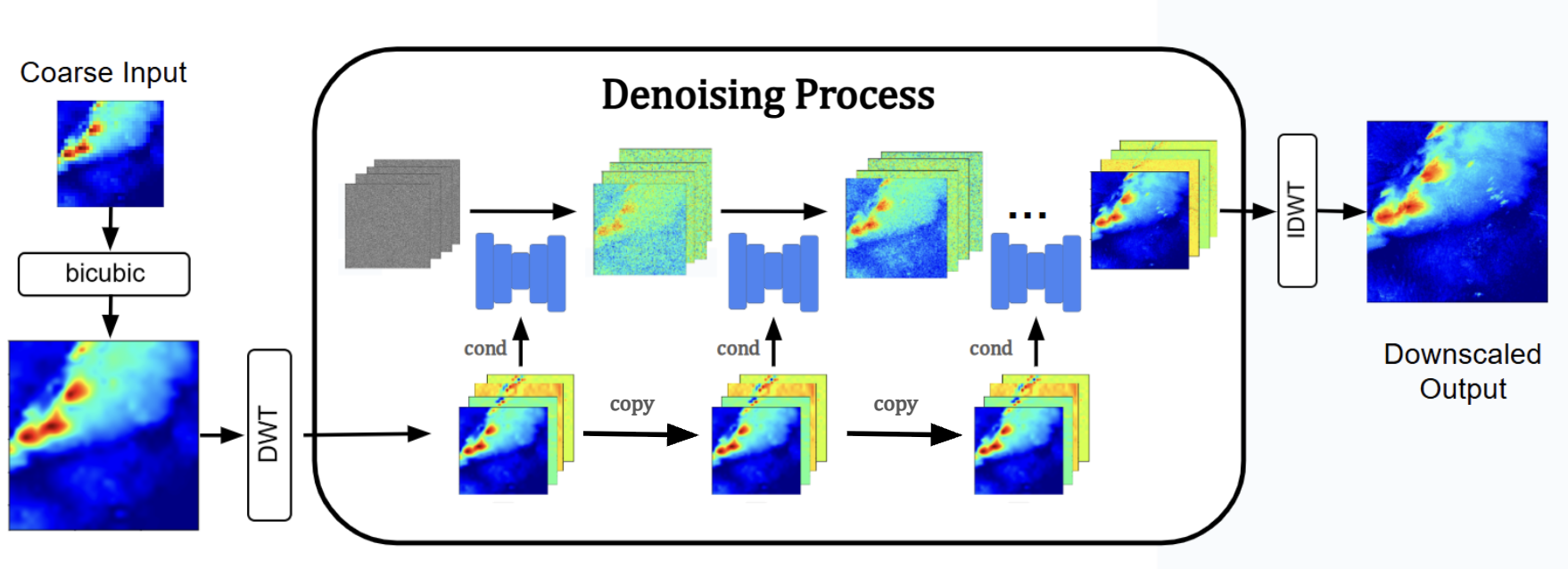}
\caption{Diagram for WDM}
\label{fig:diagram}
\end{figure}

\subsection{Wavelet Decomposition}
\label{sec:wavelet_decomp}

To effectively model the complex multi-scale structures inherent in precipitation fields, we leverage the Discrete Wavelet Transform (DWT). Meteorological phenomena such as precipitation are characterized by a wide range of spatial patterns, from large-scale frontal systems (low-frequency components) to localized, intense convective cells (high-frequency components). By wavelet transform, our model can learn the coarse structure and fine details of the precipitation field in a more efficient way, which is a significant advantage over models that operate only in the pixel domain.

The 2D DWT, denoted by $\mathcal{W}$, is an invertible transformation that decomposes an image into these frequency components. For a high-resolution image $\mathbf{x}_0^{\text{high}} \in \mathbb{R}^{H \times W}$, a single-level DWT yields four sub-bands of coefficients, each of size $\frac{H}{2} \times \frac{W}{2}$: one low-frequency approximation sub-band ($A^{\text{high}}$) and three high-frequency detail sub-bands capturing vertical ($V^{\text{high}}$), horizontal ($H^{\text{high}}$), and diagonal ($D^{\text{high}}$) features. We concatenate these sub-bands along the channel dimension to form a single coefficient tensor:
\begin{equation*}
\mathbf{c}_0^{\text{high}} = \mathcal{W}(\mathbf{x}_0^{\text{high}}) \in \mathbb{R}^{\frac{H}{2} \times \frac{W}{2} \times 4}.
\label{eq:dwt}
\end{equation*}
Similarly, the low-resolution conditional input $\mathbf{x}^{\text{low}}$ is transformed into its wavelet representation, $\mathbf{c}^{\text{low}} = \mathcal{W}(\mathbf{x}^{\text{low}})$, to provide conditioning information within the same domain. A key property of the DWT is that it is a lossless transformation, and the original image can be perfectly recovered from its wavelet coefficients by applying the Inverse DWT, denoted as $\mathcal{W}^{-1}$.

\subsection{Wavelet Diffusion Model}
\label{sec:WDM}
Our proposed Wavelet Diffusion Model (WDM) applies the conditional diffusion process directly to the wavelet coefficients defined in Section \ref{sec:wavelet_decomp}. In WDM, the forward process (Equation~\eqref{eq:forward-sde}) transforms the initial clean coefficients $\mathbf{c}_0^{\text{high}}$ into Gaussian noise $\mathbf{c}_T^{\text{high}}$, and the reverse process (Equation~\eqref{eq:reverse-sde-general}) then denoise $\mathbf{c}_T^{\text{high}}$ back to $\hat{\mathbf{c}}_0^{\text{high}}$ after learning the conditional score function $S_{\theta}(\mathbf{c}_t^{\text{high}}, \mathbf{c}^{\text{low}}, t) \approx \nabla_{\mathbf{c}} \log p_t(\mathbf{c}_t^{\text{high}} \mid \mathbf{c}^{\text{low}})$. In a similar vein, the primary training objective of WDM is the conditional score-matching loss adapted for the wavelet domain:
\begin{equation}
L_{\mathrm{cond}}(\theta) := \mathbb{E}_{t} \big\{w(t) \mathbb{E}_{(\mathbf{c}_0^{\text{high}}, \mathbf{c}^{\text{low}})} \mathbb{E}_{\mathbf{c}_{t}^{\text{high}}} \big [ \bigl\| S_{\theta}(\mathbf{c}_{t}^{\text{high}}, \mathbf{c}^{\text{low}}, t) - \nabla_{\mathbf{c}} \log p_t(\mathbf{c}_{t}^{\text{high}} \mid \mathbf{c}_{0}^{\text{high}}) \bigr\|_2^2 \big ] \big \},
\label{eq:WDM_loss}
\end{equation}
where $\mathbf{c}^{\text{high}}_{0} = \mathcal{W}(\mathbf{x}_0^{\text{high}})$ and $\mathbf{c}^{\text{low}} = \mathcal{W}(\mathbf{x}^{\text{low}})$.

While effective, relying solely on the primary loss term, $L_{\mathrm{cond}}$, can result in unwanted noise or artifacts. This issue is particularly prominent in the detail channels $(V, H, D)$, which not only capture important edge and texture information but are also where random noise predominantly resides. For radar data, this noise often manifests as unrealistic speckles or chaotic, pixel-to-pixel fluctuations in dBZ values, which leads to significant errors in meteorological products \cite{argenti2013tutorial, lee2017polarimetric, goodman1976some, lee1994speckle}. Models optimized only to minimize a pixel-wise loss will attempt to reproduce these artifacts faithfully, resulting in meteorological incoherence. To address this, we introduce a Total Variation (TV) regularization term \cite{osher2005iterative}, $R$, applied specifically to the detail coefficients ($V, H, D$), to suppress speckle while preserving important edge-like features. To compute this regularization during training without a full reverse sampling loop, we obtain an estimate of the clean coefficients, $\hat{\mathbf{c}}_0^{\text{high}} = \{\hat{A}^{\text{high}}, \hat{V}^{\text{high}}, \hat{H}^{\text{high}}, \hat{D}^{\text{high}} \}$, using Tweedie's formula \cite{efron2011tweedie}, which provides a one-step denoising prediction from step $t$.
Suppose we discretize the time interval $[0,T]$ into $N$ timesteps, 
\begin{equation}
\hat{\mathbf{c}}_0^{\text{high}}(t) \approx \frac{1}{\sqrt{\bar{\alpha}_t}}\left(\mathbf{c}^{\text{high}}_t + (1-\bar{\alpha}_t) \cdot S_{\theta}\left(\mathbf{c}^{\text{high}}_t, \mathbf{c}^{\text{low}},t\right)\right),
\label{eq:tweedie}
\end{equation}
where $\alpha_t \triangleq 1-\beta_t, \bar{\alpha}_t \triangleq \prod_{j=1}^t \alpha_j$, and $\beta_t \triangleq \beta(t T / N)$. 
The TV regularization is then imposed on the detail channels of this estimate:
\begin{equation}
R(\hat{\mathbf{c}}_0^{\text{high}}) = \text{TV}(\hat{V}^{\text{high}}) + \text{TV}(\hat{H}^{\text{high}}) + \text{TV}(\hat{D}^{\text{high}}).
\label{eq:tv_loss}
\end{equation}
The final training objective for the WDM is a weighted sum of the two losses:
\begin{equation}
L_{\text{WDM}} = L_{\mathrm{cond}} + \lambda R,
\label{eq:final_loss}
\end{equation}
where $\lambda$ is a hyperparameter balancing the two terms. 

During inference,  low-resolution observations are first transformed by $\mathbf{c}^{\text{low}} = \mathcal{W}(\mathbf{x}^{\text{low}})$. By analogy, the score network $S_{\theta}(\mathbf{c}_t, \mathbf{c}^{\text{low}}, t)$ guides the denoising process to generate wavelet coefficients for high-resolution targets $\hat{\mathbf{c}}_0^{\text{high}}$. The high-resolution precipitation field is finally obtained by applying the IDWT:
\begin{equation*}
\hat{\mathbf{x}}_0^{\text{high}} = \mathcal{W}^{-1}(\hat{\mathbf{c}}_0^{\text{high}}).
\label{eq:WDM_reconstruction}
\end{equation*}
The overall workflow of the WDM framework is depicted in Figure~\ref{fig:diagram}.

\section{Numerical Results}\label{sec:results}
\subsection{Data}\label{sec:data}
Our study utilizes the Multi-Radar Multi-Sensor (MRMS) product as the high-resolution ground truth. This dataset provides composite reflectivity at a 1-km spatial resolution by integrating data from multiple radar systems and other sources \cite{MRMSref2016}. We collected data for a region over Oklahoma, spanning longitudes $281.05\degree$E to $283.95\degree$E and latitudes $38.05\degree$N to $39.95\degree$N. The training dataset covers the period from from Jan 2, 2021 to May 20, 2021 with about 2000 samples, while data from May 6, 2024 to April 2, 2025 was reserved for testing with 160 samples. To create the corresponding low-resolution inputs for our downscaling task, we downsampled the 1-km ground truth data grid to a 10-km resolution ($30\times30$ grid) using spatial averaging, and crop the ground truth data into size $288\times288$. To focus on significant precipitation events, we first filtered the dataset according to the recorded events from the Storm Events Database \cite{NOAASED}. Specifically, the precipitation events related to convective storms (resulting in hail, tornadoes, and wind gusts) were selected. Then, we removed samples that were predominantly. Specifically, we retained only those samples where at least 50\% of the pixels in the region had non-zero reflectivity values.

\subsection{Evaluation Metrics}
We denote the generated high-resolution field as $\hat{\mathbf{x}}^{\text{high}}$ and its ground truth reference field as ${\mathbf{x}}^{\text{high}}$ , with $\hat{\mathbf{x}}^{\text{high}}[i,j]$ and ${\mathbf{x}}^{\text{high}}[i,j]$ representing the $(i,j)$ pixel, $i = 1,\dots,H$, $j = 1,\dots,W$. To evaluate the performance of our method, we employ comprehensive metrics that capture both perceptual image quality and meteorological accuracy. We define each metric in detail as follows.
\paragraph{Root Mean Squared Error (RMSE) \& Mean Absolute Error (MAE)}
RMSE and MAE are standard metrics used to quantify the average pixel-wise differences between $\hat{\mathbf{x}}^{\text{high}}$ and ${\mathbf{x}}^{\text{high}}$. They are defined as:
\begin{equation*}
\text{MAE} = \frac{1}{HW} \sum_{i=1}^{H} \sum_{j=1}^{W} |{\mathbf{x}}^{\text{high}}[i,j] - \hat{\mathbf{x}}^{\text{high}}[i,j]|,
\quad
\text{RMSE} = \sqrt{\frac{1}{HW} \sum_{i=1}^{H}\sum_{j=1}^{W}({\mathbf{x}}^{\text{high}}[i,j] - \hat{\mathbf{x}}^{\text{high}}[i,j])^2}.
\end{equation*}

\paragraph{Peak Signal-to-Noise Ratio (PSNR)}
PSNR \cite{sethi2022comprehensive} is a widely used metric for evaluating image reconstruction quality. It quantifies the ratio between the maximum possible power of a signal and the power of the corrupting noise, and higher PSNR values indicate better reconstruction quality. Here $MAX(\mathbf{x}^{\text{high}}) = 80$ represents the maximun possible values. PSNR is computed by
\begin{equation*}
\text{PSNR} = 10 \cdot \log_{10} \left( \frac{MAX(\mathbf{x}^{\text{high}})}{\frac{1}{HW}\sum_{i=1}^{H}\sum_{j=1}^{W} ( {\mathbf{x}}^{\text{high}}[i,j] - \hat{\mathbf{x}}^{\text{high}}[i,j])^2} \right).
\end{equation*}

\paragraph{Structural Similarity Index Measure (SSIM)}
SSIM \cite{4775883}, unlike pixel-wise errors, is a metric that better aligns with human visual perception by comparing the similarity of luminances ($l$), contrasts ($c$) and structures ($s$) between local patches of two images. With ${\mathbf{x}^{\text{high}}_w}$ and $\hat{\mathbf{x}}^{\text{high}}_w$ denoting the local $K \times K$ (in our case, $K = 7$) windows extracted from ${\mathbf{x}}^{\text{high}}$ and $\hat{\mathbf{x}}^{\text{high}}$ respectively, the SSIM between them is defined as:
\begin{equation*}
\begin{split}
\text{SSIM}(\mathbf{x}^{\text{high}}_w, \hat{\mathbf{x}}^{\text{high}}_w) 
&=
l(\mathbf{x}^{\text{high}}_w, \hat{\mathbf{x}}^{\text{high}}_w) \cdot c(\mathbf{x}^{\text{high}}_w, \hat{\mathbf{x}}^{\text{high}}_w) \cdot s(\mathbf{x}^{\text{high}}_w, \hat{\mathbf{x}}^{\text{high}}_w) \\ 
& = \left( \frac{2\mu_{\mathbf{x}^{\text{high}}_w}\mu_{\hat{\mathbf{x}}^{\text{high}}_w} + C_1}{\mu_{\mathbf{x}^{\text{high}}_w}^2 + \mu_{\hat{\mathbf{x}}^{\text{high}}_w}^2 + C_1} \right) 
\cdot \left( \frac{2\sigma_{\mathbf{x}^{\text{high}}_w}\sigma_{\hat{\mathbf{x}}^{\text{high}}_w} + C_2}{\sigma_{\mathbf{x}^{\text{high}}_w}^2 + \sigma_{\hat{\mathbf{x}}^{\text{high}}_w}^2 + C_2} \right)
\cdot \left( \frac{\sigma_{\mathbf{x}^{\text{high}}_w\hat{\mathbf{x}}^{\text{high}}_w} + C_3}{\sigma_{\mathbf{x}^{\text{high}}_w}\sigma_{\hat{\mathbf{x}}^{\text{high}}_w} + C_3} \right),
\end{split}
\end{equation*}

where $\mu_{\mathbf{x}^{\text{high}}_w}$ and $\mu_{\hat{\mathbf{x}}^{\text{w}}}$ are the local window means, $\sigma_{\mathbf{x}^{\text{high}}_w}$ and $\sigma_{\hat{\mathbf{x}}^{\text{high}}_w}$ are the local window standard deviation, and $\sigma_{{\mathbf{x}^{\text{high}}_w}{\hat{\mathbf{x}}^{\text{high}}_w}}$ is the covariance between $\mathbf{x}^{\text{high}}_w$ and $\hat{\mathbf{x}}^{\text{high}}_w$. $C_1$, $C_2$ and $C_3$ are small constants to avoid numerical instability during division. The final SSIM score between the entire image ${\mathbf{x}}^{\text{high}}$ and $\hat{\mathbf{x}}^{\text{high}}$ is computed by sliding the $K \times K$ window across the image pixel by pixel and averaging all the local SSIM scores. Its value ranges from -1 to 1, with 1 signifying perfect similarity in spatial patterns.

\paragraph{Critical Success Index (CSI)}
The CSI \cite{lin2005precipitation}, also known as the Threat Score (TS), is a standard meteorological metric that quantifies a model's skill in predicting an event based on a given intensity threshold. Given a reflectivity intensity threshold $\tau$, let $H$ be the number of Hits (where $\hat{\mathbf{x}}^{\text{high}}[i,j] > \tau$ and ${\mathbf{x}}^{\text{high}}[i,j] > \tau$), $M$ be the number of Misses (where $\mathbf{x}^{\text{high}}[i,j] > \tau$ and $\hat{\mathbf{x}}^{\text{high}}[i,j] \le \tau$), and $F$ be the number of False Alarms (where $\mathbf{x}^{\text{high}}[i,j] \le \tau$ and $\hat{\mathbf{x}}^{\text{high}}[i,j] > \tau$). CSI is then computed by
\begin{equation*}
\text{CSI} = \frac{H}{H + M + F},
\end{equation*}
which ranges from 0 to 1, with 1 indicating perfect prediction skill.
\paragraph{High-Intensity MSE (HI-MSE)}
This metric isolates model performance on extreme precipitation (represented by high reflectivity) by computing the Mean Squared Error only over pixels where the ground truth intensity exceeds a high threshold $\tau$. Let $\mathcal{I}_{\tau} = \{(i,j) \mid \mathbf{x}^{\text{high}}[i,j] > \tau\}$ be the set of indices for high-intensity pixels, and let $N_{\tau} = |\mathcal{I}_{\tau}|$ be the number of such pixels. HI-MSE is computed by:
\begin{equation*}
\text{HI-MSE} = \frac{1}{N_{\tau}} \sum_{(i,j) \in \mathcal{I}_{\tau}} (\mathbf{x}^{\text{high}}[i,j] - \hat{\mathbf{x}}^{\text{high}}[i,j])^2 .
\end{equation*}
Here we choose $\tau = 55$. 

\subsection{Benchmarks}
We implement the following methods and compare their performances:
\begin{enumerate}
    \item \textbf{Bicubic} \cite{opencv_library}: An bicubic interpolation method from \textit{OpenCV} package provides upsampling.
    \item \textbf{U-Net} \cite{ronneberger2015u}: A standard U-Net trained to directly regress high-resolution fields from low-resolution inputs.
    \item \textbf{EDM} \cite{karras2022elucidating}: A score-based conditional diffusion model that operates in pixel space.
    \item \textbf{FourierDM}: A conditional diffusion model that performs in the frequency domain. 
    \item \textbf{WDM}: Our proposed Wavelet Diffusion Model.
    
\end{enumerate}

\subsection{Quantitative Downscaling Performance}

We first evaluate the downscaling performance for different metrics in Table~\ref{tab:overall_performance}. The proposed WDM achieves the best performance across nearly all metrics. It produces the highest PSNR, as well as the lowest RMSE and MAE, indicating superior pixel-level accuracy. Additionally, its top performances in SSIM and spectral error reflects the preservation of multi-scale spatial patterns.
Furthermore, WDM yields the highest CSI and HRMSE, demonstrating accurate event reproduction across different intensities. 

Overall, the quantitative evaluation presented in Table~\ref{tab:overall_performance} provides strong evidence for the effectiveness of the proposed WDM framework for the task of high-resolution precipitation downscaling, outperforming both deterministic baselines and standard pixel-based diffusion models, especially in accurately reproducing spatial structures and statistical properties.

\begin{table}[h!]
\centering
\caption{Quantitative comparison of downscaling performance across different models.  Arrows indicate preferred direction (↑ higher is better, ↓ lower is better). Best and second best performance for each metric is highlighted in bold and underline.}
\label{tab:overall_performance}
\begin{tabular}{lcccccc}
\toprule
Metric(Averaged) & Bicubic & U-Net & EDM & FourierDM & WDM  \\
\midrule
MAE ↓ &3.397  &2.582  &\underline{2.372}  &  {2.475}& \textbf{2.081} \\
RMSE ↓ &5.546  &3.827  &3.446  &  \underline{3.429}& \textbf{3.127} \\
PSNR ↑ &23.430  &26.537  & 27.387 & \underline{27.422}& \textbf{28.243} \\

SSIM ↑ &\underline{0.518}  &0.474  & 0.469 & 0.459& \textbf{0.556} \\
HRMSE ↑ &283.417 & 60.130 &  \underline{58.754} & 73.357 & \textbf{53.626} \\
CSI ↑ & 0.396 & 0.586 & \underline{0.611} & 0.587 & \textbf{0.620} \\
\bottomrule
\end{tabular}
\end{table}

\begin{figure}[h!]
    \centering
    \begin{subfigure}[b]{0.48\textwidth}
        \centering
        \includegraphics[width=\textwidth]{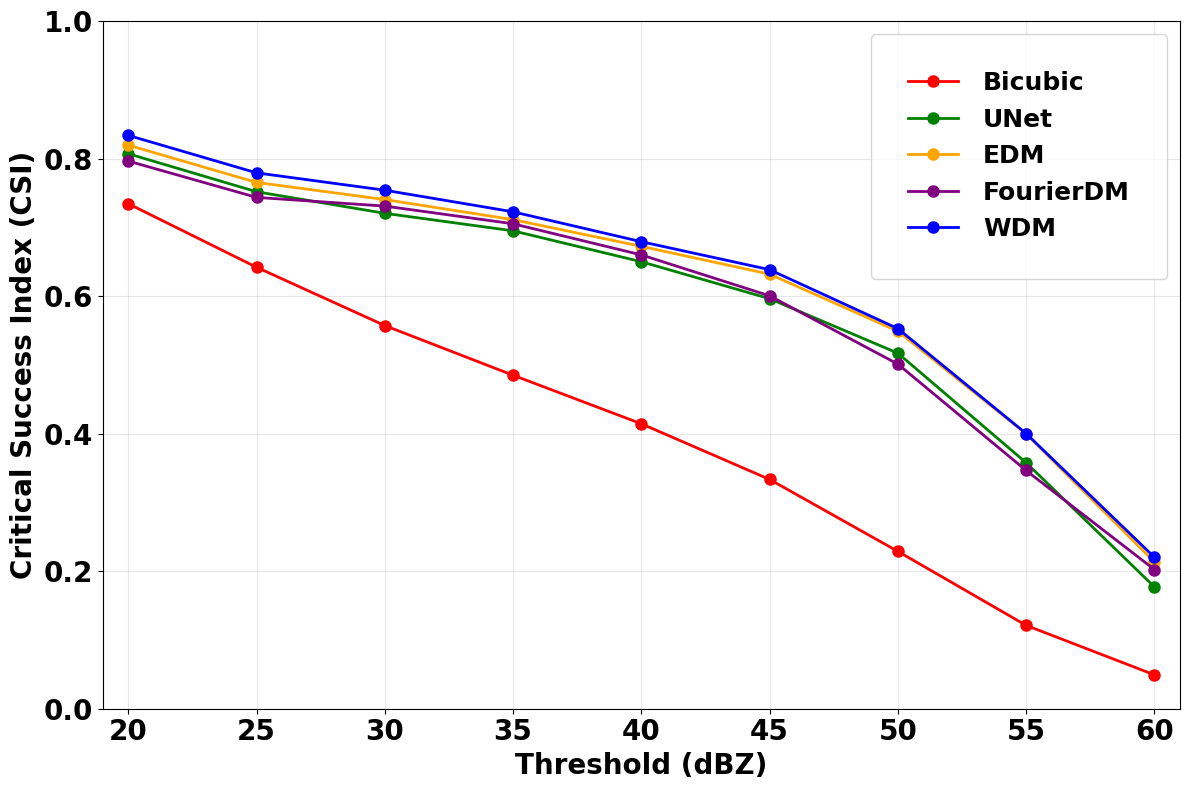}
        \label{fig:csi_thresholds}
    \end{subfigure}
    \hfill
    \begin{subfigure}[b]{0.48\textwidth}
        \centering
        \includegraphics[width=\textwidth]{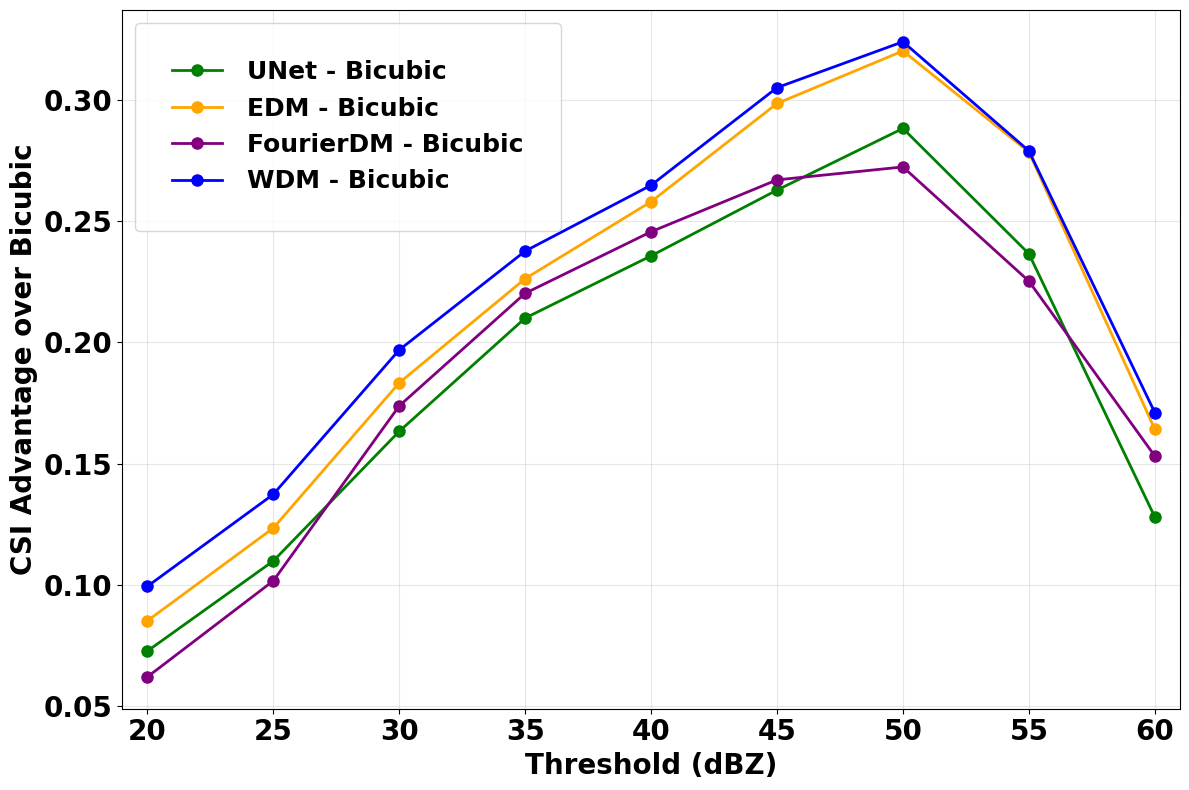}
        \label{fig:csi_difference}
    \end{subfigure}
    \caption{\textbf{Left}: CSI for different thresholds. \textbf{Right}: CSI advantages for different thresholds over bicubic method. }
    \label{fig:csi}
\end{figure}

The CSI show that WDM achieves the highest spatial accuracy across nearly all intensity thresholds, particularly for moderate and high precipitation events. We also show results of CSI for thresholds in $[20, 25,30,35,40,45,50,55,60]$ from different methods, and compute the advantage of other methods over Bicubic interpolation in Fig \ref{fig:csi}. We present the average CSI across all thresholds in Table \ref{tab:overall_performance}.

\begin{figure}

\centering
\includegraphics[width = \textwidth]{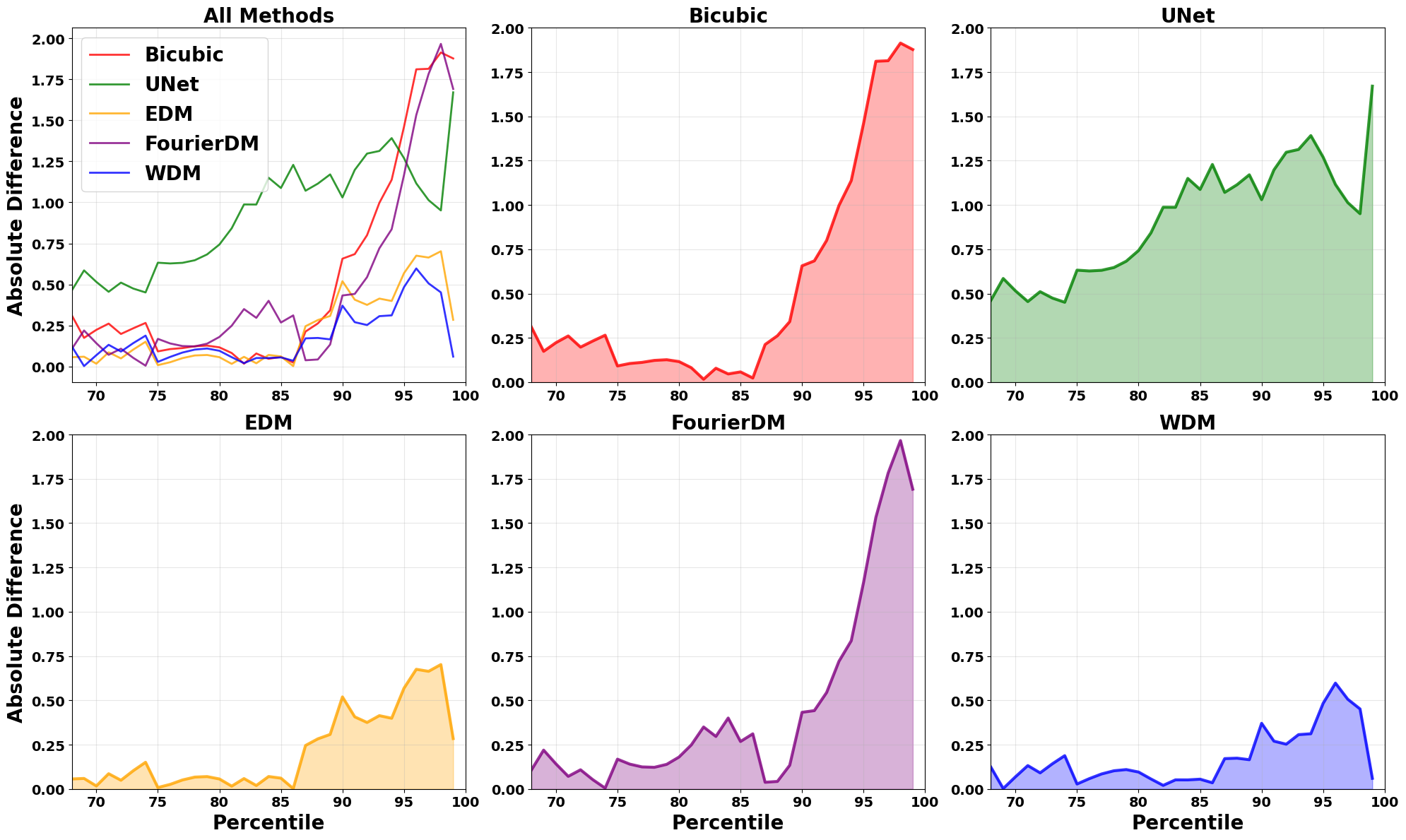}

\caption{Quantile absolute error of different methods. }
\label{fig:qplots}
\end{figure}
We also conducted a distributional analysis of model errors, which provides a more comprehensive performance assessment than standard aggregate metrics such as Mean Absolute Error (MAE) or Root Mean Square Error (RMSE). For reflectivity data, we are more concerned with extreme events when reflectivity is large, which corresponds to the higher quantiles of the distribution. We calculate the corresponding percentiles for 20, 40, 50 (dBZ), and present the absolute quantile difference over these quantiles for different methods between the ground truth in Figure~\ref{fig:qplots} and Table \ref{tab:quantile}. We observe that our WDM model is much better at fitting the overall distribution than other approaches.  

\begin{table}[h!]
\centering
\caption{Mean absolute error for the quantile for different range, where the corresponding percentiles for 20, 40, 50 (dBZ) are 68.3 \% 93.6\% and 97.5\%. Lower quantile errors indicate better distribution preservation.}
\label{tab:quantile}
\begin{tabular}{lcccccc}
\toprule
Quantile Range & Bicubic & U-Net &EDM& FourierDM & WDM  \\
\midrule
$68.3\% \sim 93.6\%$  &0.258 & 0.855 &  0.149 &0.250 & \textbf{0.127}   \\
$93.6 \% \sim 97.5 \%$ & 1.578 & 1.185 & 0.588 &1.353 & \textbf{0.485} \\
$> 97.5 \%$ & 1.922 & 1.140&  0.600& 1.903& \textbf{0.334} \\
\bottomrule
\end{tabular}
\end{table}

\subsection{Qualitative Visual Comparison}

\begin{figure}[h!]
\centering
\includegraphics[width = \textwidth,height=\textheight]{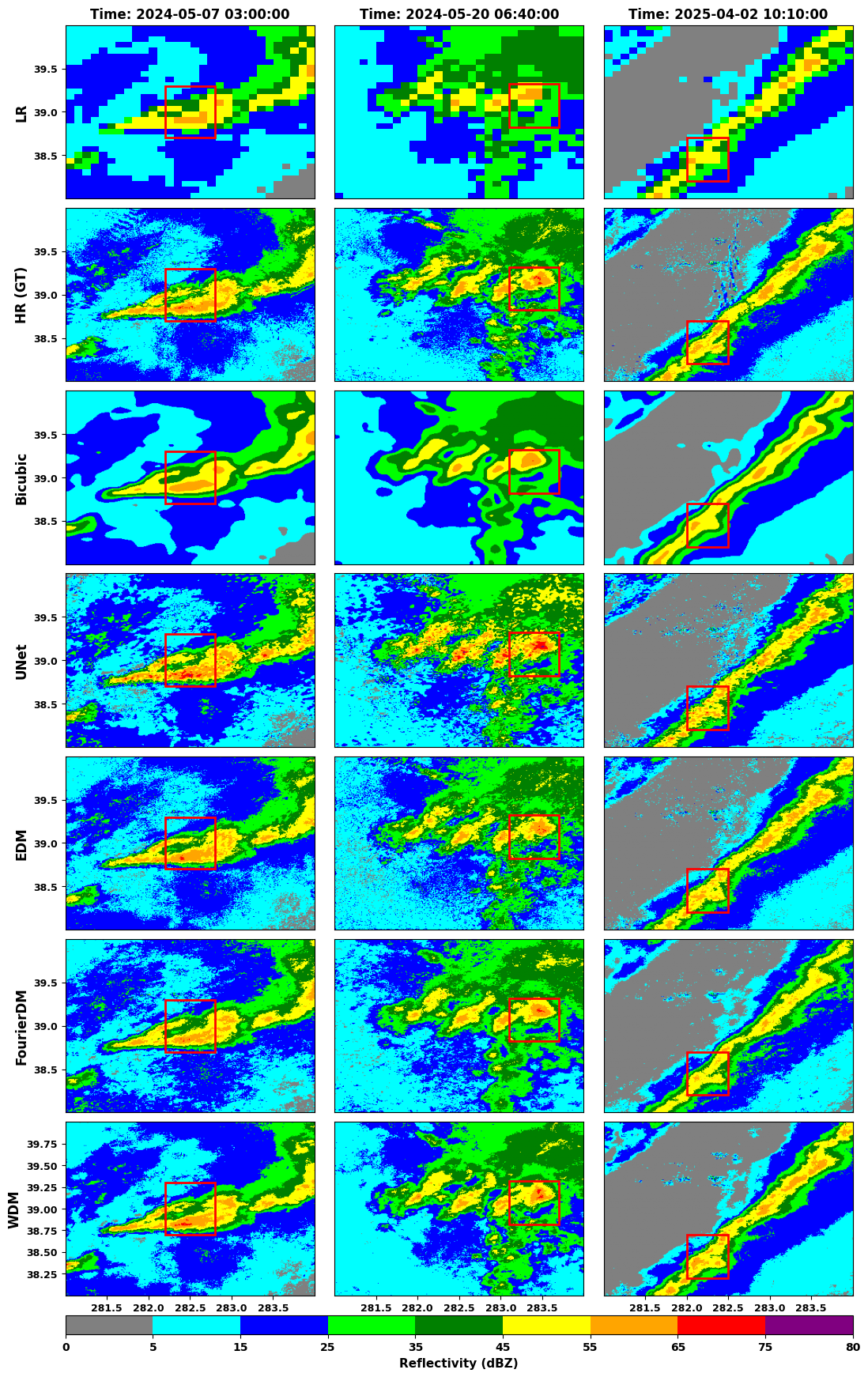}
\caption{Visual comparison of downscaling results for different representative reflectivity events. }
\label{fig:visual_comparison}
\end{figure}

Figure~\ref{fig:visual_comparison} provides a qualitative comparison of downscaled fields from representative examples to visually illustrate the effectiveness of WDM. As expected, bicubic interpolation produces an overly smooth output that fails to capture fine-scale reflectivity details. While the deterministic U-Net improves the rendering of large-scale structures, it also introduces spurious artifacts not present in the ground truth.

The diffusion-based methods generally achieve more detailed reconstructions. However, both the pixel-space EDM and FourierDM exhibit granular artifacts, particularly in regions of high reflectivity, which undermine the reliability of their extreme events modeling. Our proposed WDM yields better visually consistent results, with more realistic textures and better preservation of the sharp edges of the reflectivity fields. A notable phenomenon is the WDM's ability to suppress artifacts, even those present in the original radar observations. We attribute this to the TV regularization term, which, by design, penalizes spurious oscillations (noise) while maintaining sharp edges and textures within the generated reflectivity fields. To support the observed visual advantages, we also provide auxiliary quantitative metrics in Table \ref{tab:visual_example_table}
for the corresponding comparisons above. WDM achieves top scores on both pixel-level and multi-scale spatial metrics, which aligns with the qualitative observations and further highlights its ability to generate realistic and high-fidelity reflectivity fields across all scales.
\begin{table}[htbp]
\centering
\small
\setlength{\tabcolsep}{6pt}
\renewcommand{\arraystretch}{1.2}
\caption{Quantitative comparison of downscaling results for Figure~\ref{fig:visual_comparison}.}
\label{tab:visual_example_table}

\textbf{Time: 2024-05-07 03:00:00}

\begin{tabular}{l|c|c|c}
\bottomrule
\textbf{Method} & \textbf{PSNR} & \textbf{RMSE} & \textbf{SSIM}\\ 
\cline{1-4}
Bicubic   & 20.248 & 6.609 & \underline{0.535}  \\
\cline{1-4}
UNet      & 24.860 & 3.886 & 0.467 \\
\cline{1-4}
EDM       & \underline{26.227} & \underline{3.320} & 0.484  \\
\cline{1-4}
FourierDM & 25.582 & 3.576 & 0.456 \\
\cline{1-4}
WDM       & \textbf{26.859} & \textbf{3.087} & \textbf{0.555} \\
\toprule
\end{tabular}
\vspace{0.8em}

\textbf{Time: 2024-05-20 06:40:00}

\begin{tabular}{l|c|c|c}
\bottomrule
\textbf{Method} & \textbf{PSNR} & \textbf{RMSE} & \textbf{SSIM}  \\ \cline{1-4}
Bicubic   & 21.249 & 6.062 & \underline{0.518}  \\ \cline{1-4}
UNet      & 22.458 & 5.275 & 0.398 \\ \cline{1-4}
EDM       & 24.107 & 4.363 & 0.404 \\ \cline{1-4}
FourierDM & \underline{24.982} & \underline{3.945} & 0.461  \\ \cline{1-4}
WDM    & \textbf{25.352} & \textbf{3.780} & \textbf{0.542} \\
\toprule
\end{tabular}
\vspace{0.8em}

\textbf{Time: 2025-04-02 10:10:00}

\begin{tabular}{l|c|c|c}
\bottomrule
\textbf{Method} & \textbf{PSNR} & \textbf{RMSE} & \textbf{SSIM}  \\ \cline{1-4}
Bicubic   & 18.747 & 7.427 & 0.546 \\ \cline{1-4}
UNet      & 25.166 & 3.548 & 0.572 \\ \cline{1-4}
EDM       & \underline{26.241} & \underline{3.134} & \underline{0.585}  \\ \cline{1-4}
FourierDM & 25.571 & 3.386 & 0.481 \\ \cline{1-4}
WDM       & \textbf{26.896} & \textbf{2.907} & \textbf{0.649}  \\
\toprule
\end{tabular}
\end{table}

\subsection{Efficiency}
Our WDM framework achieves significant speedup in inference compared to standard diffusion models, while maintaining the same size of parameters. This efficiency stems from the single-level wavelet decomposition, which transforms the model's input from a $H \times W \times 1$ tensor into a multi-channel $H/2 \times W/2 \times 4$ tensor representing its frequency sub-bands. While a single convolutional layer applied to either input format performs a theoretically equivalent number of floating-point operations (FLOPs), the practical benefit emerges from the U-Net-based architecture. The primary benefit arises from the reduced spatial dimensions ($H/2 \times W/2$) of the input tensor, such that every subsequent feature map is four times smaller. The reduction of feature map decreases the cumulative computational cost and memory footprint, as the most expensive layers of the network operate on smaller feature maps. To quantify the computational efficiency of our method, we benchmarked its inference time against competing diffusion models. All experiments were conducted on a single NVIDIA A100 GPU. For a standardized comparison, each model was tasked with generating 512 total samples using a batch size of 32. To ensure reliable measurements, the benchmark for each model was executed five times, and the average inference time was reported. As shown in Table~\ref{tab:efficiency}, our WDM framework demonstrated a significant reduction in computation time with single-level wavelet decomposition. By further applying two-level wavelet decomposition to transform the input from $H \times W \times 1$ to $H/4 \times W/4 \times 16$, we obtain an approximate 9x acceleration without compromising image quality.

\begin{table}[h!]
\centering
\caption{Comparison of sampling times for a single image across different diffusion models. WDM-1 and WDM-2 denote the single-level and two-level Wavelet Diffusion Models, respectively. }
\label{tab:efficiency}
\begin{tabular}{lcccc}
\toprule
Method & Number of Parameters & Sampling Time   \\
\midrule
EDM   & 28,316,545    & 9m 16.5s  \\
FourierDM  & 28,320,002   & 12m 41.9s \\
WDM-1 & 28,313,092 &   \underline{2m 34.4s}\\
WDM-2 &  28,347,664&   \textbf{53.4s}\\
\bottomrule
\end{tabular}
\end{table}

\section{Discussion and Conclusion}\label{sec:conclusion}
Our proposed Wavelet Diffusion Model (WDM) offers a valuable tool for enhancing a variety of geoscience products \cite{HuffmanTanNCAR:GPMIMERG:2025,Mazzoglio2019, MRMSref2016}. Beyond refining observational datasets, WDM can complement the outputs of advanced machine learning forecast models like PanGu-Weather \cite{bi2023accurate}, GenCast \cite{lam2023graphcast}, and FourCastNet \cite{10.1145/3592979.3593412}. While these global models achieve remarkable speed and skill at a $0.25\degree$ resolution, their outputs can be further improved for regional impact studies that require finer detail. WDM serves as an efficient bridge, rapidly transforming coarse global model outputs into targeted, high-resolution regional forecasts. Furthermore, by generating high-resolution regional data, WDM can be used to create kilometer-scale datasets, which can in turn be used to pre-train and improve the capabilities of these large-scale ML models \cite{Harris2022,Mardani2025,lopezgomez2024dynamicalgenerativedownscalingclimatemodel}. This demonstrates the wide applicability of WDM in both observation-based and model-based workflows.

While our results are promising, we recognize several avenues for future improvement. The current evaluation was conducted exclusively using data from Oklahoma; future work should extend this analysis to other geographical regions, to test the model's performance on different meteorological phenomena. Similarly, the versatility of our framework could be explored by applying it to other crucial geophysical variables that require fine-scale detail, such as downscaling temperature, soil moisture, or wind speed. The current framework operates on fixed time instances, and a natural extension would be to incorporate temporal dynamics for spatio-temporal downscaling. Methodologically, our study was based Haar wavelet basis, and a comprehensive evaluation of other bases like Daubechies or Biorthogonal wavelets \cite{daubechies1992ten} could yield further performance gains for different types of geoscience data. Finally, we plan to explore techniques for accelerating the sampling process, such as Ordinary Differential Equation (ODE)-based samplers \cite{songdenoising, lu2022dpm} or parallel sampling strategies \cite{shih2023parallel}, to enhance the model's efficiency for real-world applications.

In summary, the WDM presents a significant advancement in generative downscaling by effectively unifying conditional diffusion modeling with wavelet-domain representations. Its strong empirical performance in generating high-resolution reflectivity fields highlights the potential of domain-aware generative models to play a crucial role in enhancing geophysical data products. The framework not only outperforms existing methods but also opens new possibilities for improving both observational data and prognostic model outputs across the geosciences.

\section{Acknowledgments}

C.Yi, M.Yu and H.Yang were partially supported by the US National Science Foundation under awards DMS2244988, DMS-2206333, the Office of Naval Research Award N00014-23-1-2007, and the DARPA D24AP00325-00.  Y.Wen and W.Qian thank the support from the Department of Geography, University of Florida. 

The authors utilized ChatGPT to refine the manuscript's spelling and grammar. This work is approved for public release; distribution is unlimited. The implementation in this study was significantly based on JAX templates from the \textit{Swirl-Dynamics} package, and C. Yi gratefully acknowledges its authors for making this resource available. 

\bibliographystyle{unsrt}

\bibliography{references}

\begin{thebibliography}{10}

\bibitem{adler2012estimating}
Robert~F Adler, Guojun Gu, and George~J Huffman.
\newblock Estimating climatological bias errors for the global precipitation climatology project (gpcp).
\newblock {\em Journal of Applied Meteorology and Climatology}, 51(1):84--99, 2012.

\bibitem{BehrangiWen2017}
Ali Behrangi and Yixin Wen.
\newblock On the spatial and temporal sampling errors of remotely sensed precipitation products.
\newblock {\em Remote Sensing}, 9(11), 2017.

\bibitem{behrangi2012quantification}
Ali Behrangi, Matthew Lebsock, Sun Wong, and Bjorn Lambrigtsen.
\newblock On the quantification of oceanic rainfall using spaceborne sensors.
\newblock {\em Journal of Geophysical Research: Atmospheres}, 117(D20), 2012.

\bibitem{Cerveny2007}
R.~S. Cerveny, J.~Lawrimore, R.~Edwards, and C.~Landsea.
\newblock Extreme weather records.
\newblock {\em Bulletin of the American Meteorological Society}, 88:853--860, 2007.

\bibitem{Wen2021}
Yixin Wen, Terry Schuur, Humberto Vergara, and Charles Kuster.
\newblock Effect of precipitation sampling error on flash flood monitoring and prediction: Anticipating operational rapid-update polarimetric weather radars.
\newblock {\em Journal of Hydrometeorology}, 22(7):1913--1929, 2021.

\bibitem{rs12081258}
Zhi Li, Mengye Chen, Shang Gao, Zhen Hong, Guoqiang Tang, Yixin Wen, Jonathan~J. Gourley, and Yang Hong.
\newblock Cross-examination of similarity, difference and deficiency of gauge, radar and satellite precipitation measuring uncertainties for extreme events using conventional metrics and multiplicative triple collocation.
\newblock {\em Remote Sensing}, 12(8), 2020.

\bibitem{behrangi2024comparative}
Ali Behrangi, Yang Song, George~J Huffman, and Robert~F Adler.
\newblock Comparative analysis of the latest global oceanic precipitation estimates from gpm v07 and gpcp v3. 2 products.
\newblock {\em Journal of Hydrometeorology}, 25(2):293--309, 2024.

\bibitem{schneider2013climate}
David~P Schneider, Clara Deser, John Fasullo, and Kevin~E Trenberth.
\newblock Climate data guide spurs discovery and understanding.
\newblock {\em Eos, Transactions American Geophysical Union}, 94(13):121--122, 2013.

\bibitem{HuffmanTanNCAR:GPMIMERG:2025}
George~J. Huffman and Jackson Tan.
\newblock The climate data guide: Imerg precipitation algorithm and the global precipitation measurement (gpm) mission, 2025.
\newblock National Center for Atmospheric Research (UCAR). Accessed on 2025-05-20. Available at \url{https://climatedataguide.ucar.edu/climate-data/gpm-global-precipitation-measurement-mission}.

\bibitem{Mazzoglio2019}
P.~Mazzoglio, F.~Laio, S.~Balbo, P.~Boccardo, and F.~Disabato.
\newblock Improving an extreme rainfall detection system with {GPM IMERG} data.
\newblock {\em Remote Sensing}, 11(6):677, 2019.

\bibitem{wilby1997downscaling}
Robert~L Wilby and Thomas~ML Wigley.
\newblock Downscaling general circulation model output: a review of methods and limitations.
\newblock {\em Progress in physical geography}, 21(4):530--548, 1997.

\bibitem{wilby1998statistical}
Robert~L Wilby, TML Wigley, D~Conway, PD~Jones, BC~Hewitson, J~Main, and DS~Wilks.
\newblock Statistical downscaling of general circulation model output: A comparison of methods.
\newblock {\em Water resources research}, 34(11):2995--3008, 1998.

\bibitem{tareghian2013statistical}
Reza Tareghian and Peter~F Rasmussen.
\newblock Statistical downscaling of precipitation using quantile regression.
\newblock {\em Journal of hydrology}, 487:122--135, 2013.

\bibitem{Fowler2007}
Hayley~J Fowler, Stephen Blenkinsop, and Claudia Tebaldi.
\newblock Linking climate c hange modelling to impacts studies.
\newblock {\em International Journal of Climatology}, 2007.

\bibitem{Wood2004}
Andrew~W. Wood, Lai~R. Leung, Venkataramana Sridhar, and Dennis~P. Lettenmaier.
\newblock Hydrologic implications of dynamical and statistical approaches to downscaling climate model outputs.
\newblock {\em Climatic Change}, 62:189--216, 2004.

\bibitem{Hong2014}
Song-You Hong and Masao Kanamitsu.
\newblock Dynamical downscaling: Fundamental issues from an nwp point of view and recommendations.
\newblock {\em Asia-Pacific Journal of Atmospheric Sciences}, 50:83--104, 2014.

\bibitem{Glotter2014}
Michael Glotter, Joshua Elliott, David McInerney, Neil Best, Ian Foster, and Elisabeth~J. Moyer.
\newblock Evaluating the utility of dynamical downscaling in agricultural impacts projections.
\newblock {\em Proceedings of the National Academy of Sciences}, 111(24):8776--8781, 2014.

\bibitem{Wang2021}
Jiali Wang, Zhengchun Liu, Ian Foster, Won Chang, Rajkumar Kettimuthu, and V.~Rao Kotamarthi.
\newblock Fast and accurate learned multiresolution dynamical downscaling for precipitation.
\newblock {\em Geoscientific Model Development Discussions}, pages 1--24, 2021.

\bibitem{dai2025precipdiff}
Ting-Yu Dai and Hayato Ushijima-Mwesigwa.
\newblock Precipdiff: Leveraging image diffusion models to enhance satellite-based precipitation observations.
\newblock {\em arXiv preprint arXiv:2501.07447}, 2025.

\bibitem{ledig2017photo}
Christian Ledig, Lucas Theis, Ferenc Husz{\'a}r, Jose Caballero, Andrew Cunningham, Alejandro Acosta, Andrew Aitken, Alykhan Tejani, Johannes Totz, Zehan Wang, et~al.
\newblock Photo-realistic single image super-resolution using a generative adversarial network.
\newblock In {\em Proceedings of the IEEE conference on computer vision and pattern recognition}, pages 4681--4690, 2017.

\bibitem{Harris2022}
Lucy Harris, Andrew T.~T. McRae, Matthew Chantry, Peter~D. Dueben, and Tim~N. Palmer.
\newblock A generative deep learning approach to stochastic downscaling of precipitation forecasts.
\newblock {\em Journal of Advances in Modeling Earth Systems}, 14(10):e2022MS003120, 2022.

\bibitem{lugmayr2020srflow}
Andreas Lugmayr, Martin Danelljan, Luc Van~Gool, and Radu Timofte.
\newblock Srflow: Learning the super-resolution space with normalizing flow.
\newblock In {\em Computer vision--ECCV 2020: 16th European conference, glasgow, UK, August 23--28, 2020, proceedings, part v 16}, pages 715--732. Springer, 2020.

\bibitem{ho2020denoising}
Jonathan Ho, Ajay Jain, and Pieter Abbeel.
\newblock Denoising diffusion probabilistic models.
\newblock {\em Advances in neural information processing systems}, 33:6840--6851, 2020.

\bibitem{song2020score}
Yang Song, Jascha Sohl-Dickstein, Diederik~P Kingma, Abhishek Kumar, Stefano Ermon, and Ben Poole.
\newblock Score-based generative modeling through stochastic differential equations.
\newblock {\em arXiv preprint arXiv:2011.13456}, 2020.

\bibitem{dhariwal2021diffusion}
Prafulla Dhariwal and Alexander Nichol.
\newblock Diffusion models beat gans on image synthesis.
\newblock {\em Advances in neural information processing systems}, 34:8780--8794, 2021.

\bibitem{Ling2024}
Fenghua Ling, Zeyu Lu, Jing-Jia Luo, Lei Bai, Swadhin~K. Behera, Dachao Jin, Baoxiang Pan, Huidong Jiang, and Toshio Yamagata.
\newblock Diffusion model-based probabilistic downscaling for 180-year east asian climate reconstruction.
\newblock {\em npj Climate and Atmospheric Science}, 7(1):131, 2024.

\bibitem{Harder2023}
Paula Harder, Alex Hernandez-Garcia, Venkatesh Ramesh, Qidong Yang, Prasanna Sattegeri, Daniela Szwarcman, Campbell~D. Watson, and David Rolnick.
\newblock Hard-constrained deep learning for climate downscaling.
\newblock {\em Journal of Machine Learning Research}, 24(1), 2023.

\bibitem{lopezgomez2024dynamicalgenerativedownscalingclimatemodel}
Ignacio Lopez-Gomez, Zhong~Yi Wan, Leonardo Zepeda-Núñez, Tapio Schneider, John Anderson, and Fei Sha.
\newblock Dynamical-generative downscaling of climate model ensembles.
\newblock {\em Proceedings of the National Academy of Sciences}, 122(17):e2420288122, 2025.

\bibitem{watt2024generativediffusionbaseddownscalingclimate}
Robbie~A. Watt and Laura~A. Mansfield.
\newblock Generative diffusion-based downscaling for climate, 2024.

\bibitem{Mardani2025}
M.~Mardani, N.~Brenowitz, Y.~Cohen, et~al.
\newblock Residual corrective diffusion modeling for km-scale atmospheric downscaling.
\newblock {\em Communications Earth \& Environment}, 6:124, 2025.

\bibitem{bell2021era5}
Bill Bell, Hans Hersbach, Adrian Simmons, Paul Berrisford, Per Dahlgren, Andr{\'a}s Hor{\'a}nyi, Joaqu{\'\i}n Mu{\~n}oz-Sabater, Julien Nicolas, Raluca Radu, Dinand Schepers, et~al.
\newblock The era5 global reanalysis: Preliminary extension to 1950.
\newblock {\em Quarterly Journal of the Royal Meteorological Society}, 147(741):4186--4227, 2021.

\bibitem{lussana2022wavelet}
Cristian Lussana, Barbara Casati, Rasmus~E Benestad, Julia Lutz, Andreas Dobler, Oskar Landgren, Jan~Erik Haugen, Abdelkader Mezghani, and Kajsa~M Parding.
\newblock Wavelet transform applied to era5 global daily precipitation fields to assess changes in the rainfall patterns.
\newblock Technical report, Copernicus Meetings, 2022.

\bibitem{kumar1994wavelet}
Praveen Kumar and Efi Foufoula-Georgiou.
\newblock Wavelet analysis in geophysics: An introduction.
\newblock {\em Wavelets in geophysics}, 4:1--43, 1994.

\bibitem{hersbach2020era5}
Hans Hersbach, Bill Bell, Paul Berrisford, Shoji Hirahara, Andr{\'a}s Hor{\'a}nyi, Joaqu{\'\i}n Mu{\~n}oz-Sabater, Julien Nicolas, Carole Peubey, Raluca Radu, Dinand Schepers, et~al.
\newblock The era5 global reanalysis.
\newblock {\em Quarterly journal of the royal meteorological society}, 146(730):1999--2049, 2020.

\bibitem{simmons2020global}
Adrian Simmons, Cornel Soci, Julien Nicolas, Bill Bell, Paul Berrisford, Rossana Dragani, Johannes Flemming, Leopold Haimberger, Sean Healy, Hans Hersbach, Andras Horányi, Antje Inness, J.~Munoz-Sabater, Raluca Radu, and Dinand Schepers.
\newblock Global stratospheric temperature bias and other stratospheric aspects of era5 and era5.1.
\newblock Technical report, European Centre for Medium-Range Weather Forecasts (ECMWF), 2020.
\newblock ECMWF Technical Memorandum.

\bibitem{zhang2016multi}
Jian Zhang, Kenneth Howard, Carrie Langston, Brian Kaney, Youcun Qi, Lin Tang, Heather Grams, Yadong Wang, Stephen Cocks, Steven Martinaitis, et~al.
\newblock Multi-radar multi-sensor (mrms) quantitative precipitation estimation: Initial operating capabilities.
\newblock {\em Bulletin of the American Meteorological Society}, 97(4):621--638, 2016.

\bibitem{karras2022elucidating}
Tero Karras, Miika Aittala, Timo Aila, and Samuli Laine.
\newblock Elucidating the design space of diffusion-based generative models.
\newblock {\em Advances in neural information processing systems}, 35:26565--26577, 2022.

\bibitem{liu2024generative}
Yuhao Liu, James Doss-Gollin, Guha Balakrishnan, and Ashok Veeraraghavan.
\newblock Generative precipitation downscaling using score-based diffusion with wasserstein regularization.
\newblock {\em arXiv preprint arXiv:2410.00381}, 2024.

\bibitem{phung2023wavelet}
Hao Phung, Quan Dao, and Anh Tran.
\newblock Wavelet diffusion models are fast and scalable image generators.
\newblock In {\em Proceedings of the IEEE/CVF conference on computer vision and pattern recognition}, pages 10199--10208, 2023.

\bibitem{jiang2023low}
Hai Jiang, Ao~Luo, Haoqiang Fan, Songchen Han, and Shuaicheng Liu.
\newblock Low-light image enhancement with wavelet-based diffusion models.
\newblock {\em ACM Transactions on Graphics (TOG)}, 42(6):1--14, 2023.

\bibitem{anderson1982reverse}
Brian~DO Anderson.
\newblock Reverse-time diffusion equation models.
\newblock {\em Stochastic Processes and their Applications}, 12(3):313--326, 1982.

\bibitem{vincent2011connection}
Pascal Vincent.
\newblock A connection between score matching and denoising autoencoders.
\newblock {\em Neural computation}, 23(7):1661--1674, 2011.

\bibitem{argenti2013tutorial}
Fabrizio Argenti, Alessandro Lapini, Tiziano Bianchi, and Luciano Alparone.
\newblock A tutorial on speckle reduction in synthetic aperture radar images.
\newblock {\em IEEE Geoscience and remote sensing magazine}, 1(3):6--35, 2013.

\bibitem{lee2017polarimetric}
Jong-Sen Lee and Eric Pottier.
\newblock {\em Polarimetric radar imaging: from basics to applications}.
\newblock CRC press, 2017.

\bibitem{goodman1976some}
Joseph~W Goodman.
\newblock Some fundamental properties of speckle.
\newblock {\em JOSA}, 66(11):1145--1150, 1976.

\bibitem{lee1994speckle}
Jong-Sen Lee, L~Jurkevich, Piet Dewaele, Patrick Wambacq, and Andr{\'e} Oosterlinck.
\newblock Speckle filtering of synthetic aperture radar images: A review.
\newblock {\em Remote sensing reviews}, 8(4):313--340, 1994.

\bibitem{osher2005iterative}
Stanley Osher, Martin Burger, Donald Goldfarb, Jinjun Xu, and Wotao Yin.
\newblock An iterative regularization method for total variation-based image restoration.
\newblock {\em Multiscale Modeling \& Simulation}, 4(2):460--489, 2005.

\bibitem{efron2011tweedie}
Bradley Efron.
\newblock Tweedie’s formula and selection bias.
\newblock {\em Journal of the American Statistical Association}, 106(496):1602--1614, 2011.

\bibitem{MRMSref2016}
Travis~M. Smith, Valliappa Lakshmanan, Gregory~J. Stumpf, Kiel~L. Ortega, Kurt Hondl, Karen Cooper, Kristin~M. Calhoun, Darrel~M. Kingfield, Kevin~L. Manross, Robert Toomey, and Jeff Brogden.
\newblock Multi-radar multi-sensor (mrms) severe weather and aviation products: Initial operating capabilities.
\newblock {\em Bulletin of the American Meteorological Society}, 97(9):1617 -- 1630, 2016.

\bibitem{NOAASED}
{National Centers for Environmental Information (NCEI)} and National Weather~Service (NWS).
\newblock Storm events database.
\newblock \url{https://www.ncdc.noaa.gov/stormevents/}, 2013.
\newblock Accessed: 2025-03-01.

\bibitem{sethi2022comprehensive}
Dimple Sethi, Sourabh Bharti, and Chandra Prakash.
\newblock A comprehensive survey on gait analysis: History, parameters, approaches, pose estimation, and future work.
\newblock {\em Artificial Intelligence in Medicine}, 129:102314, 2022.

\bibitem{4775883}
Zhou Wang and Alan~C. Bovik.
\newblock Mean squared error: Love it or leave it? a new look at signal fidelity measures.
\newblock {\em IEEE Signal Processing Magazine}, 26(1):98--117, 2009.

\bibitem{lin2005precipitation}
Charles Lin, Slavko Vasi{\'c}, Alamelu Kilambi, Barry Turner, and Isztar Zawadzki.
\newblock Precipitation forecast skill of numerical weather prediction models and radar nowcasts.
\newblock {\em Geophysical research letters}, 32(14), 2005.

\bibitem{opencv_library}
G.~Bradski.
\newblock {The OpenCV Library}.
\newblock {\em Dr. Dobb's Journal of Software Tools}, 2000.

\bibitem{ronneberger2015u}
Olaf Ronneberger, Philipp Fischer, and Thomas Brox.
\newblock U-net: Convolutional networks for biomedical image segmentation.
\newblock In {\em Medical image computing and computer-assisted intervention--MICCAI 2015: 18th international conference, Munich, Germany, October 5-9, 2015, proceedings, part III 18}, pages 234--241. Springer, 2015.

\bibitem{bi2023accurate}
Kaifeng Bi, Lingxi Xie, Hengheng Zhang, Xin Chen, Xiaotao Gu, and Qi~Tian.
\newblock Accurate medium-range global weather forecasting with {3D} neural networks.
\newblock {\em Nature}, 619(7970):533--538, 2023.
\newblock Breakthrough AI model surpassing ECMWF operational system.

\bibitem{lam2023graphcast}
Remi Lam, Alvaro Sanchez-Gonzalez, Matthew Willson, Peter Wirnsberger, Meire Fortunato, Alexander Pritzel, Suman Ravuri, Timo Ewalds, Ferran Alet, Zaheer Abbas, et~al.
\newblock Learning skillful medium-range global weather forecasting.
\newblock {\em Science}, 382:1416--1421, 2023.
\newblock GraphCast: AI-based global weather model surpassing traditional NWP.

\bibitem{10.1145/3592979.3593412}
Thorsten Kurth, Shashank Subramanian, Peter Harrington, Jaideep Pathak, Morteza Mardani, David Hall, Andrea Miele, Karthik Kashinath, and Anima Anandkumar.
\newblock Fourcastnet: Accelerating global high-resolution weather forecasting using adaptive fourier neural operators.
\newblock In {\em Proceedings of the Platform for Advanced Scientific Computing Conference}, PASC '23, New York, NY, USA, 2023. Association for Computing Machinery.

\bibitem{daubechies1992ten}
Ingrid Daubechies.
\newblock {\em Ten lectures on wavelets}.
\newblock SIAM, 1992.

\bibitem{songdenoising}
Jiaming Song, Chenlin Meng, and Stefano Ermon.
\newblock Denoising diffusion implicit models.
\newblock In {\em International Conference on Learning Representations}, 2021.

\bibitem{lu2022dpm}
Cheng Lu, Yuhao Zhou, Fan Bao, Jianfei Chen, Chongxuan Li, and Jun Zhu.
\newblock Dpm-solver: A fast ode solver for diffusion probabilistic model sampling in around 10 steps.
\newblock {\em Advances in Neural Information Processing Systems}, 35:5775--5787, 2022.

\bibitem{shih2023parallel}
Andy Shih, Suneel Belkhale, Stefano Ermon, Dorsa Sadigh, and Nima Anari.
\newblock Parallel sampling of diffusion models.
\newblock {\em Advances in Neural Information Processing Systems}, 36:4263--4276, 2023.

\bibitem{tancik2020fourier}
Matthew Tancik, Pratul Srinivasan, Ben Mildenhall, Sara Fridovich-Keil, Nithin Raghavan, Utkarsh Singhal, Ravi Ramamoorthi, Jonathan Barron, and Ren Ng.
\newblock Fourier features let networks learn high frequency functions in low dimensional domains.
\newblock {\em Advances in neural information processing systems}, 33:7537--7547, 2020.

\bibitem{jax2018github}
James Bradbury, Roy Frostig, Peter Hawkins, Matthew~James Johnson, Chris Leary, Dougal Maclaurin, George Necula, Adam Paszke, Jake Vander{P}las, Skye Wanderman-{M}ilne, and Qiao Zhang.
\newblock {JAX}: composable transformations of {P}ython+{N}um{P}y programs, 2018.
\newblock Available at \url{http://github.com/jax-ml/jax}.

\bibitem{flax2020github}
Jonathan Heek, Anselm Levskaya, Avital Oliver, Marvin Ritter, Bertrand Rondepierre, Andreas Steiner, and Marc van {Z}ee.
\newblock {F}lax: A neural network library and ecosystem for {JAX}, 2024.
\newblock Available at \url{http://github.com/google/flax}.

\bibitem{deepmind2020jax}
DeepMind, Igor Babuschkin, Kate Baumli, Alison Bell, Surya Bhupatiraju, Jake Bruce, Peter Buchlovsky, David Budden, Trevor Cai, Aidan Clark, Ivo Danihelka, Antoine Dedieu, Claudio Fantacci, Jonathan Godwin, Chris Jones, Ross Hemsley, Tom Hennigan, Matteo Hessel, Shaobo Hou, Steven Kapturowski, Thomas Keck, Iurii Kemaev, Michael King, Markus Kunesch, Lena Martens, Hamza Merzic, Vladimir Mikulik, Tamara Norman, George Papamakarios, John Quan, Roman Ring, Francisco Ruiz, Alvaro Sanchez, Laurent Sartran, Rosalia Schneider, Eren Sezener, Stephen Spencer, Srivatsan Srinivasan, Milo\v{s} Stanojevi\'{c}, Wojciech Stokowiec, Luyu Wang, Guangyao Zhou, and Fabio Viola.
\newblock The {D}eep{M}ind {JAX} {E}cosystem, 2020.
\newblock http://github.com/google-deepmind.

\bibitem{loshchilov2017decoupled}
Ilya Loshchilov and Frank Hutter.
\newblock Decoupled weight decay regularization.
\newblock {\em arXiv preprint arXiv:1711.05101}, 2017.

\bibitem{haar1910theorie}
Alfr{\'e}d Haar.
\newblock {\em Zur theorie der orthogonalen functionensysteme. inaugural}.
\newblock PhD thesis, 1910.

\bibitem{wolter2021frequency}
Moritz Wolter.
\newblock {\em Frequency domain methods in recurrent neural networks for sequential data processing}.
\newblock PhD thesis, Universit{\"a}ts-und Landesbibliothek Bonn, 2021.

\end{thebibliography}

\appendix

\section{Implementation Details}
\label{sec:implementation}
\subsection{Experimental Setup}
All models use the same U-Net \cite{ronneberger2015u} backbone for fair comparison. The low-resolution input $\mathbf{x}^{\text{low}}$  is first upsampled using bicubic interpolation and concatenated channel-wise with the input $\mathbf{x}_t^{\text{high}}$. The encoder includes multiple downsampling stages with progressively increasing channels (96, 192, 384), each consisting of two convolutional residual blocks. Each stage employs two convolutional residual blocks. Self-attention layers are added at the bottleneck (coarsest resolution) before beginning the upsampling decoder path. In diffusion models (EDM, FourierDM and WDM), the diffusion timestep 
$t$ is encoded using Fourier features \cite{ tancik2020fourier} and injected into residual blocks via affine transformations. We implement our codes in JAX \cite{jax2018github}, with Flax \cite{flax2020github} for neural architectures and Optax  \cite{deepmind2020jax} for optimization. All models were trained using the Adam optimizer \cite{loshchilov2017decoupled} with $1\times10^{-4}$ stepsize. An exponential moving average (EMA) of the model weights was maintained with a decay factor of 0.999. All models were trained for 1000 epochs with batch size 32 in a single A100 GPU. Specifically, we choose $\lambda=10^{-5}$ in training WDM.

\subsection{Diffusion Model Implementation}
The objective in diffusion models can be parameterized in several equivalent ways. While our methodology section introduces the framework via the score function $S_{\theta}$, for practical implementation we adopt the preconditioned denoising framework proposed in EDM \cite{karras2022elucidating}. The network is parameterized as a denoiser, $D_\theta(\mathbf{c}^{\text{high}}_\sigma, \mathbf{c}^{\text{low}}, \sigma)$, which aims to approximate the clean coefficients $\mathbf{c}_0^{\text{high}}$. Here, the network's goal is to directly predict the clean data from a noised version. This denoiser $D_\theta$ is mathematically equivalent to score-prediction network $S_{\theta}$ \cite{song2020score}:
\begin{equation*}
    S_{\theta}(\mathbf{c}_\sigma, \mathbf{c}^{\text{low}},\sigma) \approx \frac{D_\theta(\mathbf{c}^{\text{high}}_\sigma, \mathbf{c}^{\text{low}}, \sigma) - \mathbf{c}^{\text{high}}_\sigma}{\sigma^2}
\end{equation*}
Here we replace $t$ with $\sigma$ for the noise level follows \cite{karras2022elucidating}. More specifically, the drift $f(t) = -\frac{1}{2}\beta(t)$ and diffusion $g(t)=\sqrt{\beta(t)}$ coefficients in \cite{karras2022elucidating} are reparametrized by
\begin{equation*}
s(t) = \exp (\int_0^t f(\xi) d\xi) \text{ and } \sigma(t) = \sqrt{\int_0^t \frac{g^2(\xi)}{s^2(\xi)} d\xi}.
\end{equation*} 
In our implementation, we first preprocess the data by normalizing it to have zero mean and unit variance, and set scaling schedule in \cite{karras2022elucidating} as $s(t) = \frac{1}{\sqrt{1+\sigma^2(t)}}$ and $\sigma(t)$ with cosine schedule \cite{dhariwal2021diffusion}. We sample $p_\sigma(\sigma)$ uniformly sampled in $t$. 
This formulation provides a better performance in practice. The specific choices for noise loss weighting $w(\sigma)$ in training are critical hyperparameters, and we follow the recommendations provided in Table 1 in \cite{karras2022elucidating}. For solving the SDE (sampling), we use the Euler-Maruyama method with 300 steps and employ Equation (5) in \cite{karras2022elucidating}.

\subsection{Wavelet Transform Implementation}
A central component of WDM is the use of a single-level Discrete Wavelet Transform (DWT) with a Haar basis \cite{haar1910theorie}, implemented via the \textit{jaxwt} \cite{wolter2021frequency} package. Specifically, applying DWT to $\mathbf{x}_0^{\text{high}}$ results in a four-channel tensor $\mathbf{c}_{0}$ with spatial dimensions of $144 \times 144$. The DWT maps the 288×288 image into a four-channel 144×144 tensor, representing low- and high-frequency components $(A, V, H, D)$. Conditions on coarse images are first upscaled by bicubic interpolation to $288 \times 288$ such that its wavelet coefficients have the same size as our target resolution. The wavelet coefficients of interpolated conditions are then concatenated channel-wise to the network input directly. In the inference stage, after the reverse SDE generates the denoised wavelet coefficients $\hat{\mathbf{c}}_0$ during inference, the final high-resolution image $\hat{\mathbf{x}}_0^{\text{high}}$ ($288 \times 288$) is reconstructed using the corresponding Inverse DWT (IDWT), $\mathcal{W}^{-1}$.

\end{document}